# Surrogate-assisted cooperative signal optimization for large-scale traffic networks

Yongsheng Liang[a], Zhigang Ren[a], Lin Wang[b], Hanqing Liu[a], and Wenhao Du[a]

[a] School of Automation Science and Engineering, Xi'an Jiaotong University, Xi'an, China
[b] School of Information Science and Technology, Northwest University, Xi'an, China

**Corresponding author:** Zhigang Ren

**Affiliation:** School of Automation Science and Engineering, Xi'an Jiaotong University

**Address:** No.28 Xianning West Road, Xi'an, Shaanxi, 710049, P.R. China

**Email address:** renzg@mail.xjtu.edu.cn

# Surrogate-assisted cooperative signal optimization for large-scale traffic networks

Yongsheng Liang[a], Zhigang Ren[a], Lin Wang[b], Hanqing Liu[a], and Wenhao Du[a]

[a] School of Automation Science and Engineering, Xi'an Jiaotong University, Xi'an, China

[b] School of Information Science and Technology, Northwest University, Xi'an, China

**Abstract:**

Reasonable setting of traffic signals can be very helpful in alleviating congestion in urban traffic networks. Meta-heuristic optimization algorithms have proved themselves to be able to find high-quality signal timing plans. However, they generally suffer from performance deterioration when solving large-scale traffic signal optimization problems due to the huge search space and limited computational budget. Directing against this issue, this study proposes a surrogate-assisted cooperative signal optimization (SCSO) method. Different from existing methods that directly deal with the entire traffic network, SCSO first decomposes it into a set of tractable sub-networks, and then achieves signal setting by cooperatively optimizing these sub-networks with a surrogate-assisted optimizer. The decomposition operation significantly narrows the search space of the whole traffic network, and the surrogate-assisted optimizer greatly lowers the computational burden by reducing the number of expensive traffic simulations. By taking Newman fast algorithm, radial basis function and a modified estimation of distribution algorithm as decomposer, surrogate model and optimizer, respectively, this study develops a concrete SCSO algorithm. To evaluate its effectiveness and efficiency, a large-scale traffic network involving crossroads and T-junctions is generated based on a real traffic network. Comparison with several existing meta-heuristic algorithms specially designed for traffic signal optimization demonstrates the superiority of SCSO in reducing the average delay time of vehicles.

## 1. Introduction

Traffic congestion has always been a serious issue in many cities around the world, resulting in economic loss, air pollution and inconvenience to residents. Improving traffic capacity by extending the current traffic network is not always viable due to the high cost. Transportation authorities and researchers are interested in making better use of the current traffic infrastructures with intelligent traffic signal management methods [1,2]. A proper schedule of traffic signals can be very helpful to improve the traffic flow and reduce the delay time of vehicles.

Current research efforts in traffic signal management methods can be basically categorized into two classes, i.e. fixed-time methods [3,4] and traffic responsive methods [5,6]. The former aims to find suitable fixed-time signal plans based on historical traffic demand, while the latter dynamically adjusts the signal state according to the traffic information detected in real time. Although traffic responsive methods are technically sound, their performance depends heavily on real-time sensor systems [7] and they are generally difficult to apply to the whole city owing to the high operational cost [8,9]. Besides, the majority of traffic lights in real-world work under fixed signal timing plans and the traffic flows tend to repeat similar patterns like morning and evening peaks. Therefore, developing efficient fixed-time traffic signal optimization methods is of great practical significance.

Traffic signal management has attracted lots of research attention in the intelligent transportation domain, and various approaches, such as deep learning [10], fuzzy systems [11] and meta-heuristic algorithms [1-4,12,13], have been applied in this domain. Among them, meta-heuristic algorithms, such as genetic algorithm (GA) [14] and particle swarm optimization (PSO) [4], show certain superiority. They have been successfully employed to tackle small and medium-scale traffic signal optimization problems, but have seldom been applied in large-scale traffic networks.

The challenges of meta-heuristic algorithms in solving large-scale traffic signal optimization problems mainly lie in two aspects. On the one hand, with the growth of the number of traffic signals, the solution space of the signal optimization problem increases exponentially and the traffic flows in the network also become more complex. As a result, basic meta-heuristic algorithms can hardly find satisfying signal plans. This is the so-called "curse of dimensionality" effect [15]. On the other hand, the signal optimization process relies on a traffic simulation model to simulate the traffic flow dynamics and thus to evaluate the effectiveness of candidate signal plans, the process of which is very computationally expensive [16,17]. The situation becomes more serious with the increase of the network scale, and only a small number of simulations could be allowed [1]. This greatly limits the performance of meta-heuristic algorithms as they generally require many simulations to evolve their individuals before finding a satisfying solution.

Large-scale traffic signal optimization problem is a typical large-scale and computationally expensive optimization problem. To tackle such kind of problems, the recently developed surrogate-assisted cooperative co-evolution (SACC) method [18] demonstrates great potential. By taking the idea of "divide-and-conquer", SACC first decomposes the original large-scale problem into a set of smaller sub-problems such that the whole search space can be greatly reduced and a surrogate model can be easily trained for each sub-problem. The surrogate model helps the optimizer to approximately and efficiently evaluate candidate solutions of the corresponding sub-problem and thus to locate high-quality sub-solutions. By cooperatively optimizing these sub-problems with the surrogate-assisted optimizer, SACC outperforms traditional optimizers in solving large-scale and expensive optimization problems. Despite its appealing performance, SACC was merely tested on some benchmark functions and has never been applied to deal with real-world problems.

Aiming at solving large-scale traffic signal optimization problem and also bridging the gap between algorithm research and engineering practice, this study proposes a surrogate-assisted cooperative signal optimization (SCSO) algorithm based on the key idea of SACC. The main contributions of this study are summarized as follows:

1) A surrogate-assisted cooperative signal optimization framework is designed for large-scale traffic signal optimization. SCSO first decomposes a large-scale traffic network into a set of smaller sub-networks, and then achieves signal optimization of the original large-scale network by cooperatively optimizing the signal plans of these sub-networks with a surrogate-assisted optimizer. As for the cooperation, different sub-networks share their current best signal plans with each other such that some promising candidate signal plans of the sub-network in hand can be formally simulated by the traffic simulator and a surrogate model can be efficiently established for this sub-network.

2) This study employs Newman fast algorithm [19] to divide a large-scale traffic network into smaller and tractable sub-networks. According to the computationally expensive characteristic of the traffic signal optimization problem, a surrogate-assisted optimizer named RBF-EDA$^2$ is developed by taking radial basis function (RBF) [20] as the surrogate model and a modified estimation of distribution algorithm (EDA) called EDA$^2$ [21] as the optimizer. Profiting from the capability of RBF in reducing the number of expensive traffic simulations and the powerful optimization ability of EDA$^2$, RBF-EDA$^2$ is capable of finding high-quality signal plans for sub-networks with tight computational budget.

3) With the goal of reducing the average delay time of vehicles in a large-scale traffic network, this study develops a concrete SCSO algorithm by integrating Newman fast algorithm and RBF-EDA$^2$ into the SCSO framework. The performance of SCSO is systematically evaluated in a large-scale traffic network involving 43 junctions generated based on a real traffic network. Crossroads and T-junctions in this network are designated with different signal phases to provide a more realistic simulation environment, and morning and evening peak hours are considered to test the performance of SCSO in different traffic situations. Comparison with several existing meta-heuristic algorithms for traffic signal optimization indicates that SCSO could have an edge over its competitors in reducing the average delay time of vehicles.

The remainder of this paper is structured as follows. Section 2 introduces the backgrounds of traffic signal optimization and surrogate-assisted cooperative co-evolution. Section 3 describes the details of the proposed SCSO algorithm. Case study is presented in Section 4 including experimental setup and result analysis. Conclusions and some future research topics are finally given in Section 5.

## 2. Background

### 2.1 Traffic signal optimization

Traffic signals play an essential role in regulating traffic flows in urban traffic networks. Fixed-time traffic signal optimization methods devote to find proper fixed signal timing plans to improve the service capacities of existing traffic networks. Based on different actual demands, various optimization objectives can be adopted in the traffic signal optimization process [14], such as improving the total traffic flow or global mean speed of vehicles, reducing the average delay time of vehicles or total gas emission, etc. Because of that it is impracticable to test the effectiveness of candidate signal plans in real traffic networks, researches generally couple a traffic simulation model with an optimization algorithm to deal with the traffic signal optimization problem.

Traffic simulation model works as an evaluation tool to assess the effectiveness of candidate signal plans generated by the optimization algorithm. Lots of traffic simulation models have been proposed from different perspectives and they can be generally classified into three types including macroscopic models, mesoscopic models and microscopic models [1,22]. Macroscopic traffic models [23] treat traffic flow as a flow of media (such as fluids or gases) and describe the macro-level behaviors of traffic flow (like density and velocity) using similar physical equations. Microscopic traffic models [17,24,25] directly take each individual vehicle as the basic unit and present the traffic dynamics of vehicle-to-vehicle interactions. Mesoscopic models [26] absorb the characteristics of the previous two types of models. With the fast development of computer simulation techniques, microscopic models are becoming more and more popular since it could generate detailed simulation results and provide deep insight into the traffic dynamics [27,28]. But the attractive performance of microscopic models also brings heavy computational burden, especially for large-scale traffic networks.

Traffic signal optimization has attracted increasing research efforts since the birth of traffic light. Earlier signal optimization methods like the canonical Webster method [29] were developed to provide proper signal timing plan for an isolated junction, establishing the basis for modern traffic optimization methods. With the rapid expansion of metropolis, more and more traffic lights are installed and the traffic flows are also becoming more complex. The continuous upsurge of traffic demands motivated the development of new technologies.

The complexity of traffic signal optimization problem has been preventing the use of traditional analytic methods. Mate-heuristic algorithms have played an important role in the advances achieved in traffic signal optimization domain. Mate-heuristic algorithms often take their inspirations from natural phenomena such as species evolution and intelligent behavior of animals [1,12,13]. These characteristics endow mate-heuristic algorithms with strong optimization ability and flexibility in complex environments. In the last two decades, various mate-heuristics have been employed to tackle the traffic signal optimization problem. Among them, GA and PSO are perhaps the two most popular algorithms. Rouphail et al. [30] attempted to optimize the traffic signal of an urban traffic network with nine intersections using a classic GA, with the aim of reducing the link delay and total network queue time. Sánchez-Medina et al. [14] developed a signal optimization platform by combining a GA with cellular automata-based microscopic simulator. The whole optimization platform is implemented on a Beowulf Cluster multicomputer to achieve better scalability and enhanced efficiency. A traffic network with 7 intersections was employed to test the performance of the optimization platform with several separate optimization objectives such as reducing the travel time and gas emission. Kachroudi and Bhouri [31] combined a multi-objective PSO with model predictive control to optimize two conflicting objectives in a 16-intersection virtual urban network. Hu et al. [4] suggested a signal timing scheduling (TSO) algorithm based on a quantum PSO. They mapped an urban traffic network with 15 junctions into an enhanced Biham, Middleton and Levine traffic model to test the performance of TSO. In addition to GA and PSO, Jovanović [32] utilized a bee colony optimization algorithm to optimize the signal setting of a traffic network with nine intersections. Experimental results showed that the employed algorithm indicated better performance than the simulated annealing algorithm in reducing the travel time of vehicles. Liang et al. [33] introduced two modified estimation of distribution algorithms for traffic signal optimization to reduce the total vehicle delay time. Experiments on a traffic network with 11 intersections indicated that the two modified EDAs could outperform several existing mate-heuristic algorithms. Osorio and Bierlaire [34] proposed a surrogate-assisted signal optimization algorithm by embedding a surrogate model within a trust region algorithm. The surrogate model works as a computationally cheap alternative to the microscopic traffic simulation model and could assist the trust region algorithm in locating promising signal plans, which will be further reevaluated by the traffic simulation model and used to update the surrogate model. Empirical analyses on two traffic networks with up to 15 signalized intersections showed that the surrogate-assisted algorithm could effectively reduce the average travel time of vehicles with decreased number of traffic simulations.

The above signal optimization algorithms showed appealing performance in small and medium-scale traffic networks.

However, signal optimization of large-scale traffic networks is still very challenging and only a few related works were presented. Garcia-Nieto et al. [7] employed a modified PSO algorithm to tackle the large-scale traffic signal optimization problem, and showed that the proposed algorithm can find appealing signal plan for traffic networks involving up to 40 junctions. Lately, Gao et al. [3] comprehensively investigated the effectiveness of five algorithms, including GA, artificial bee colony (ABC) algorithm, discrete harmony search (DHS) algorithm, Jaya algorithm and water cycle algorithm, on a set of traffic networks involving 9 to 400 intersections with the objective of reducing the total delay time of vehicles.

A common characteristic of the above methods is that they all directly optimize the signal plan of the whole traffic network with an optimization algorithm. With the increase of the number of traffic signals, the solution space of the signal optimization problem will grow rapidly. Consequently, it is very hard to obtain satisfying results by directly tackling the entire traffic network. As for surrogate models, although they are useful to reduce the expensive traffic simulations during the signal optimization process in medium-scale networks, it is however very difficult to build accurate surrogate models for a large-scale traffic network. To achieve area-wide traffic signal management, an effective approach is to divide a large-scale traffic network into a set of smaller sub-networks and then cooperatively control the traffic signal of these sub-networks. A number of network decomposition algorithms have been proposed from different perspectives to achieve reasonable traffic decomposition results [19,35-37]. For instance, Newman fast algorithm [19,36] attempts to decompose a large-scale network by maximizing the network modularity. Existing traffic network decomposition algorithms are generally combined with control methods for traffic management [38-40]. For example, Zhou et al. [38] developed a two-level hierarchical model predictive control framework for large-scale traffic network, in which the lower-level control focuses on managing the traffic signal of each sub-network and the upper-level control aims to balance the traffic flows among different sub-networks. Ding et al. [39] attempted to improve the traffic network performance by controlling the boundaries of sub-networks with a state transfer risk decision perimeter control method. These successful research works evidence that decomposing a large-scale traffic network into smaller sub-networks can significantly reduce the problem complexity and improve the traffic management efficiency. Nevertheless, traffic network decomposition methods have seldom been combined with mate-heuristic algorithms to realize cooperative signal optimization. This leaves much room for further research.

## 2.2 Surrogate model assisted cooperative co-evolution

Surrogate model assisted cooperative co-evolution (SACC) [18] is a recently developed variant of cooperative co-evolution (CC) [41,42], it improves CC by using surrogate mode to assist the sub-problem optimization process. CC is a powerful tool for solving large-scale optimization problems. It first decomposes a large-scale optimization problem into a set of tractable sub-problems and then cooperatively optimizes these sub-problems with a traditional optimization algorithm. Due to the black-box feature of optimization problem, the produced sub-problems generally do not have clear objective function or simulation model. The evaluation of a sub-solution requires cooperation with other sub-solutions. CC achieves indirect evaluation of a sub-solution by inserting it into a complete solution called context vector. And different sub-problems cooperate with each other by sharing their representative sub-solutions to constitute the context vector.

The context-vector-based evaluation method provides a practicable way to optimize sub-problems, it also greatly restricts the efficiency of traditional CC algorithms since this evaluation method totally relies on the original high-dimensional simulation model to evaluate sub-solutions in the optimization process. As a result, traditional CC algorithms often require an enormous amount of expensive simulations to obtain a relatively good solution. To address this issue, SACC constructs and maintains a surrogate model for each sub-problem. When optimizing a sub-problem, the candidate sub-solutions produced at each generation are first evaluated by the trained surrogate model. Only a part of high-quality sub-solutions refined by the surrogate model will be re-evaluated by the original simulation model, which are in turn used to update the surrogate model in the next generation. In this way, the accuracy of surrogate model could be gradually improved to identify better sub-solutions. SACC reduces the computational cost to a great extent without significantly sacrificing the optimization performance, providing an efficient framework for solving large-scale optimization problems. But the performance of SACC was only verified on a set of benchmark functions. Experimental results indicated that SACC is capable of solving different kinds of complex high-dimensional functions, but it still needs to consume considerable number of simulations before obtaining desirable results. The reason mainly lies in that the surrogate-assisted optimizer in SACC needs to evaluate a certain percentage (like 10%) of sub-solutions with the original simulation model at each generation. Considering the huge simulation requirement of the original CC, the consumption of simulations in SACC is still relatively high.

Since first proposed in 1994, CC has got extensive research attention from different fields due to its appealing performance. Most existing related works mainly focus on developing new CC algorithms and testing their performance on some benchmark functions [43,44]. By contrast, the applications of CC in real-world optimization problems are relatively few. In the domain of intelligent transportation, CC algorithms have been applied to tackle routing problems [45,46] and conflict avoidance problem [47], but have not been employed to solve traffic signal optimization problem. As for the newly developed SACC, it has never been applied to practical problems up to now. Besides, the total number of simulations required by SACC is also unacceptable for many real-world expensive optimization problems like traffic signal optimization problem. The above considerations motivate us to develop more efficient and suitable algorithm for large-scale traffic signal optimization problem based on the key idea of SACC.

## 3. The proposed method

This section first introduces the traffic optimization problem and the solution encoding scheme, and then presents the surrogate-assisted cooperative signal optimization framework. After that, the surrogate-assisted optimizer therein named RBF-EDA$^2$ is described in detail. By integrating Newman fast decomposition algorithm and RBF-EDA$^2$ into the SCSO framework, a concrete SCSO algorithm is finally developed.

### 3.1 Problem description and solution encoding

A traffic network generally contains three basic components, i.e. roads, junctions and traffic lights. Traffic lights installed at each junction control the traffic flows in different roads. Most traffic lights in real-world traffic networks work according to predetermined fixed-time signal cycles. Traffic lights at the same junction are ruled by a common signal cycle, while lights at different junctions are usually controlled by different signal cycles. A signal cycle is a repetition of the basic series of signal phases at a junction, and each signal phase is a signal state (green, red, etc.) combination of traffic lights, enabling compatible vehicles to safely cross the corresponding junction.

To control traffic flows in the whole network, traffic lights at different junctions concurrently repeat their respective signal cycles. The purpose of traffic signal optimization is to find a set of proper signal cycles for junctions such that certain objectives, such as minimizing the average delay time of vehicles or total gas emission, can be achieved. The delay time of a vehicle is the time difference between its real travel time and ideal travel time. Drivers generally hope to travel through a local area as soon as possible, so minimizing the average delay time of vehicles is of great significance in the sense of improving drivers' satisfaction degree and alleviating traffic congestion. By means of modern microscopic traffic simulators such as VISSIM and SUMO, researchers can set up a virtual traffic network and directly simulate the effectiveness of a specified signal plan. As a result, the average delay time of vehicles can be collected and employed as the fitness value of the signal plan.

Existing research works mainly focus on optimizing signals for crossroads [3,4,33] and less attention has been paid to T-junctions, although the latter is also very common in real-world traffic networks. It would be more reasonable and practical if different types of junctions are taken into account during the signal optimization process. This study considers signal phase configurations for both crossroads and T-junctions. Figs. 1 and 2 diagrammatize the corresponding signal phases.

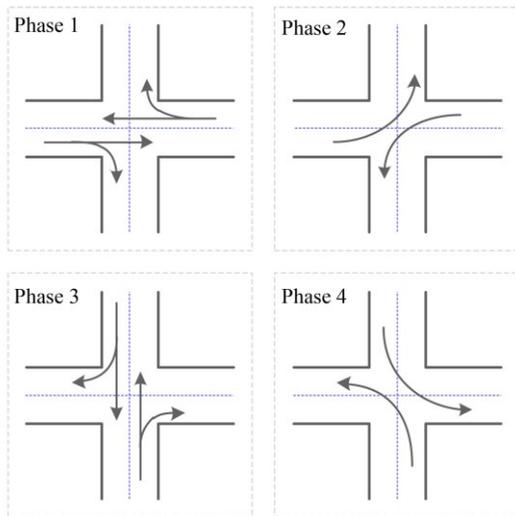
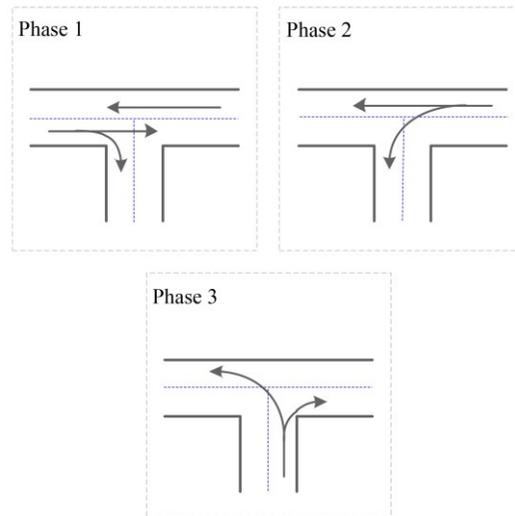

Figure 1. Four signal phases of a signal cycle for crossroads.

Figure 2. Three signal phases of a signal cycle for T-junctions.

Considering that the purpose of signal optimization is to find suitable signal cycles for all junctions and each signal cycle is composed of several signal phases, we encode a signal plan into a vector of integers, where each integer represents a signal phase duration time in a specified signal cycle and is limited to a certain range, ensuring that vehicles in different roads can safely cross a junction within reasonable time. To make it more clear, a simple solution encoding example is presented in Table 1. Three junctions are considered in this example, among which the first two are crossroads and the third one is a T-junction. As there are four (or three) phases in the signal cycle for a crossroad (or T-junction), we employ an integer vector of 11 dimensions to encode the signal plan for this simple example. This means that, for a general traffic signal optimization problem, the length of each of its candidate solution, i.e., the problem dimension *d*, equals the total number of signal phases in all junctions. For a traffic network with dozens of junctions, the dimension of the corresponding signal optimization problem will generally exceed 100, which is considered large-scale for existing optimization field. In addition, since the evaluation of a signal plan using microscopic traffic simulator is computationally expensive, the total number of available traffic simulations for signal optimization is very limited, which further increases the solving difficulty of the large-scale signal optimization problem.

Table 1. Solution encoding for two crossroads and a T-junction.

|  | Crossroad | | | | Crossroad | | | | T-Junction | | |
|---|---|---|---|---|---|---|---|---|---|---|---|
| Phase | 1 | 2 | 3 | 4 | 1 | 2 | 3 | 4 | 1 | 2 | 3 |
| Duration (s) | 40 | 33 | 42 | 26 | 32 | 38 | 25 | 40 | 39 | 37 | 28 |
| Solution | (40, 33, 42, 26, 32, 38, 25, 40, 39, 37, 28) | | | | | | | | | | |

## 3.2 Surrogate-assisted cooperative signal optimization algorithm

### 3.2.1 Framework of SCSO

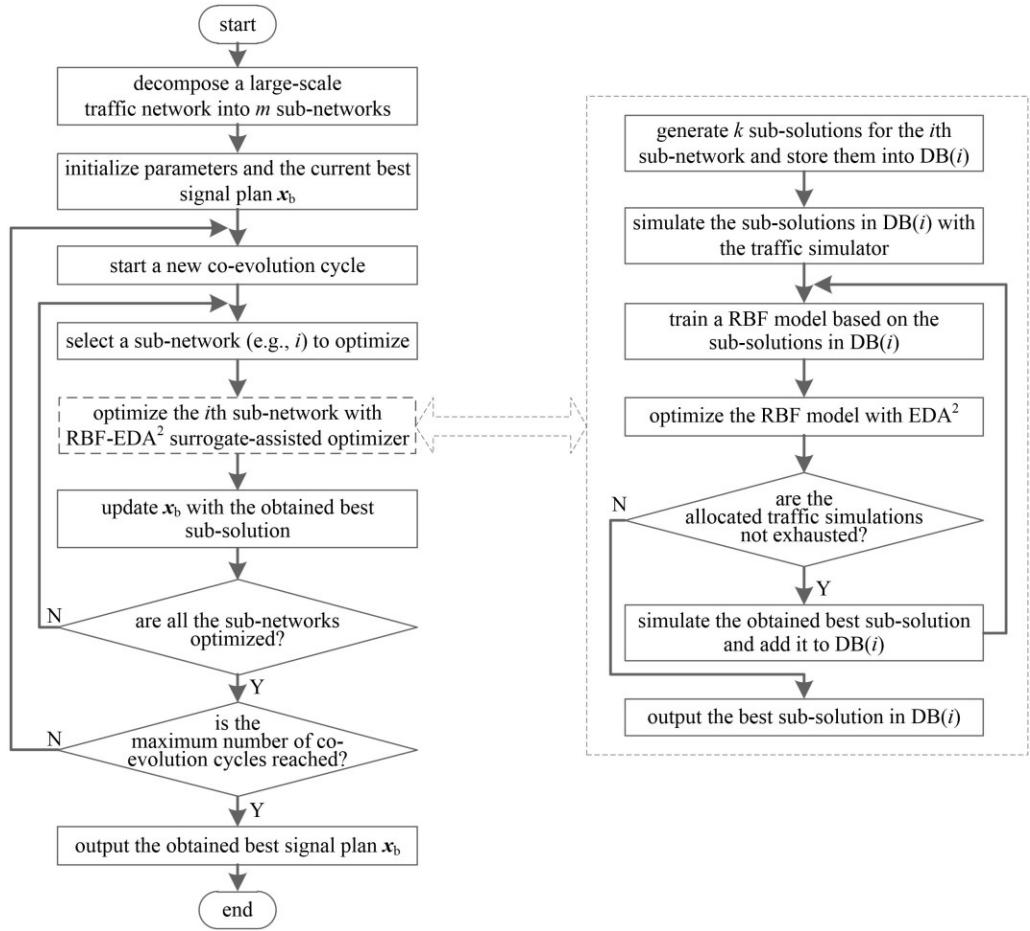

Figure 3. The framework of SCSO and the surrogate-assisted optimizer RBF-EDA$^2$.

To efficiently tackle the large-scale traffic signal optimization problem, this study proposes a surrogate-assisted cooperative signal optimization algorithm inspired by the idea of SACC. Figure 3 presents the framework of SCSO. Before the optimization process, SCSO first decomposes a large-scale traffic network into a set of smaller sub-networks such that the whole search space can be reduced. Then it initializes parameters and the current best signal plan for the whole network, which will be employed to assist in simulating the signal plan of each sub-network. Next, SCSO divides the entire optimization process into several co-evolution cycles according to a predetermined co-evolution cycle number $c$. In each co-evolution cycle, SCSO sequentially optimizes the signal plan of each sub-network with a surrogate-assisted optimizer. After all the sub-networks being tackled, SCSO starts a new co-evolution cycle to further improve the whole signal plan until the maximum number of co-evolution cycles is reached.

SCSO achieves area-wide coordinate signal optimization by simulating the signal plan of each sub-network with the help of the current complete best signal plan. To be specific, it partitions a complete solution $x$ (signal plan for the entire network) into $m$ sub-solutions $x = (x(1), x(2), ..., x(i), ..., x(m))$ according to the network decomposition result, where sub-solution $x(i)$ represents the signal plan for the $i$th sub-network. Besides, it maintains the current complete best solution $x_b = (x(1)_b, x(2)_b, ..., x(i)_b, ..., x(m)_b)$, which is composed of the current best sub-solutions of all sub-networks. When simulating a new sub-solution $x(i)_{new}$, SCSO first inserts it into the corresponding positions in $x_b$ and thus obtains a new complete solution $x_b(x(i)_{new}) = (x(1)_b, x(2)_b, ..., x(i)_{new}, ..., x(m)_b)$. Then the simulation of sub-solution $x(i)_{new}$ can be indirectly achieved by simulating $x_b(x(i)_{new})$ with the traffic simulator for the whole network. Once the optimization process of a sub-network is completed, the current complete best solution $x_b$ is immediately updated by inserting the obtained best sub-solution into it and then the updated $x_b$ will be further used to assist in simulating signal plan of other sub-networks. In this way, information exchange among different sub-networks is achieved by sharing $x_b$ in the simulation process, which contributes to improve the signal coordination level and thus can help SCSO find more proper signal plan for the entire traffic network.

According to the framework of SCSO, it can be known that network decomposition and sub-network signal optimization are two key parts of SCSO. As mentioned in the literature review section, many traffic network decomposition algorithms have been successfully proposed from different perspectives. But they are generally integrated into some signal control methods and have seldom been combined with mate-heuristic algorithms to achieve cooperative signal optimization. So in this work, we directly employ a famous decomposition algorithm, i.e. the Newman fast algorithm [19], to decompose a large-scale traffic network into sub-networks. Newman fast algorithm attempts to find proper network decomposition results by maximizing the network modularity, which have been shown to be effective in partitioning traffic networks [36]. Next we mainly focus on describing the proposed surrogate-assisted optimizer RBF-EDA$^2$ for SCSO.

### 3.2.2 RBF-EDA$^2$

Since traffic simulation is very computationally expensive, it is impractical to simulate each signal plan for sub-networks during the cooperative optimization process. Directing against this issue, we develop a surrogate-assisted optimizer for

SCSO. It takes the well-known RBF and a modified EDA called $EDA^2$ as surrogate model and optimizer, respectively, and is thus named $RBF-EDA^2$.

The right part of Fig. 3 depicts the flowchart of $RBF-EDA^2$. To optimize the signal plan of a sub-network, $RBF-EDA^2$ first randomly generates a set of sub-solutions and maintains them in a database DB. These sub-solutions will be simulated by the traffic simulator and then a RBF model is trained for this sub-network based on these simulated sub-solutions. The trained RBF model can be seen as an approximation of the objective function for the corresponding sub-network, which is very useful to locate promising signal plans. Therefore, SCSO conducts signal optimization with $EDA^2$ for a certain number of iterations merely based on the trained RBF model. This means that all the candidate sub-solutions generated by $EDA^2$ are evaluated by the RBF model, which greatly reduces the simulation requirement. After a specified number of iterations, the best sub-solution obtained by $EDA^2$ will be re-evaluated by the traffic simulator if traffic simulations allocated to this sub-network are not exhausted. This simulated sub-solution will be added to the database and further employed to update the RBF model, leading to a more accurate model that helps to find better new sub-solutions. By this means, $RBF-EDA^2$ achieves a good trade-off between optimization performance and computational efficiency.

To be specific, when optimizing the $i$th sub-network, $k$ sub-solutions $x(i)_1, x(i)_2,..., x(i)_k$ are first randomly generated and stored in a database DB($i$). These sub-solutions are simulated by the traffic simulator with the help of the current best solution $x_b$ as described in section 3.2.1. Then a RBF model can be built based on the sub-solutions in DB($i$) for the current sub-network. A modified RBF model suggested in [20] is adopted because it is relatively easy to train and is robust to different problem dimensions [48,49]. For a new sub-solution $x(i)_{new}$, its objective value can be approximately estimated by the RBF model as follows:

$$\overline{f}(x(i)_{new}) = \sum_{j=1}^{k} \omega_j \phi(\|x(i)_{new} - x(i)_j\|) + \beta^T x(i)_{new} + \alpha, \tag{1}$$

where $\phi(\cdot)$ is a basis function, $\|x(i)_{new} - x(i)_j\|$ represents the Euclidean distance between the two sub-solutions, $\omega, \beta$ and $\alpha$ are parameters. $\beta^T x(i)_{new} + \alpha$ is a polynomial tail added to the standard RBF model, which is beneficial to improve the approximation accuracy. As suggested in [20], cubic function is adopted as the basis function $\phi(\cdot)$. The detailed training method of this RBF model can be conveniently accessed in [18,20] and will not be repeated here for saving space. According to the suggestions in [20,49], $k$ is set to three times of the sub-problem dimension.

The trained RBF can be viewed as a computationally efficient approximation model for a sub-network and is employed to estimate objective values of candidate sub-solutions generated by $EDA^2$ [21]. $EDA^2$ is a recently developed variant of Gaussian estimation of distribution algorithm (GEDA). It inherits the advantages of GEDA in capturing variable dependencies with a Gaussian probability model and further improves GEDA by exploiting the evolution direction information hidden in historical solutions. For traffic signal optimization, $EDA^2$ is able to reveal the relationships among different signal phases and achieve coordinate signal optimization to some extent in complicated traffic situations. Despite its appealing performance, $EDA^2$ has never been used in traffic signal optimization.

$EDA^2$ employs a Gaussian probability model to describe the distribution of high-quality solutions and to generate new

solutions. The Gaussian model can be parameterized by a mean $\mu$ and a covariance matrix $\Sigma$ as follows:

$$G_{(\mu, \Sigma)}(x) = \frac{(2\pi)^{-n/2}}{(\det \Sigma)^{1/2}} \exp(-(x-\mu)^{T}(\Sigma)^{-1}(x-\mu)/2), \tag{2}$$

Note that, for the convenience of description, here we use complete solution $x$ rather than sub-solution $x(i)$ to formulate the Gaussian model of EDA². For a new generation, EDA² estimates its mean using the traditional maximum likelihood estimation method based on some promising solutions selected from the current population:

$$\bar{\mu}^{t+1} = \frac{1}{|S^{t}|} \sum_{i=1}^{|S^{t}|} S_{i}^{t}, \tag{3}$$

where $S_{i}^{t}$ denotes the $i$th solution in the selected solution set $S^{t}$ in the $t$th generation. Once the new mean $\bar{\mu}^{t+1}$ is determined, EDA² estimates the new covariance matrix as follows:

$$\bar{\Sigma}^{t+1} = \frac{1}{|H^{t}|} \sum_{i=1}^{|H^{t}|} (H_{i}^{t} - \bar{\mu}^{t+1})(H_{i}^{t} - \bar{\mu}^{t+1})^{T}, \tag{4}$$

where $H^{t} = S^{t} \cup S^{t-1} \cup S^{t-2} \cup ... \cup S^{t-l}$ is an archive and stores historical good solutions selected in previous $l$ generations besides the ones in $S^{t}$. By replacing $S^{t}$ with $H^{t}$, EDA² naturally integrates the evolution information of the population into the estimated Gaussian model, which endows it with a more proper search direction, a larger search scope, and thus a stronger search ability. Moreover, the utilization of historical solutions in the estimation method also contributes to reducing the population size of EDA², leading to a better convergence ability.

The detailed steps of EDA² can be directly found in [21]. According to the suggestions therein, the population size and archive length $l$ of EDA² are set to 200 and 10, respectively, in this study. As shown in Fig. 3, RBF-EDA² may invoke EDA² several times in each run. We do not require EDA² to find the globally optimal sub-solution at each time. Instead, a locally optimal sub-solution is also acceptable as it also reflects the solution space characteristics of the traffic signal optimization problem. With the accumulation of locally optimal sub-solutions, more accurate RBF models could be built, which in turn can help EDA² to locate better sub-solutions. Based on the above considerations and some preliminary experimental analyses, we set the maximum number of iterations for each run of EDA² to 100, which is sufficient for it to converge to a promising solution.

### 3.2.3 Procedure of SCSO

By taking Newman fast algorithm and RBF-EDA² as decomposer and sub-network optimizer, respectively, a concrete SCSO algorithm is developed. The whole procedure of SCSO is shown in Algorithm 1. SCSO first initializes the number of co-evolution cycles $c$ and the current complete best solution $x_b$, it should be noted that $x_b$ is initialized at the midpoint of the feasible solution region in order to provide a stable basis for the following signal optimization process. After decomposing a large-scale traffic network into $m$ sub-networks using Newman fast algorithm, the computational budget, i.e. the allowed maximum number of traffic simulations (Max_TS), is evenly allocated to each sub-network in each co-evolution cycle. That is to say, the number of available traffic simulations for RBF-EDA² at each time will be Max_TS/($c \cdot m$). Besides, a simple rounding function is also employed in step 11 to ensure that the elements, i.e. signal phase duration times, in the new sub-solution are all integers.

**Algorithm 1**: Procedure of SCSO
---
1. Initialize the number of co-evolution cycles $c$ and the current complete best solution $x_b$;
2. Decompose a traffic network into $m$ sub-networks using Newman fast algorithm;
3. Evenly allocate the computational budget;
4. **for** $j = 1$ to $c$
5.     **for** $i = 1$ to $m$
6.         Randomly generate $k$ sub-solutions and store them into DB($i$);
7.         Simulate the sub-solutions in DB($i$) using the traffic simulator with the help of $x_b$;
8.         **while** *the allocated traffic simulations are not exhausted* **do**
9.             Build a RBF model based on the sub-solutions in DB($i$);
10.            Conduct signal optimization with EDA$^2$ for a certain number of iterations based on the trained RBF model, and obtain a new sub-solution $x(i)_{new}$;
11.            Set $x(i)_{new} = Rounding(x(i)_{new})$;
12.            Simulate $x(i)_{new}$ using the traffic simulator by inserting it into $x_b$, and add $x(i)_{new}$ to DB($i$);
13.         **end while**
14.         Update $x_b$ with the best sub-solution in DB($i$);
15.     **end for**
16. **end for**
17. Output $x_b$.
---

The proposed surrogate-assisted cooperative signal optimization algorithm is not a simple application of SACC, some beneficial attempts have been made to improve its signal optimization ability in large-scale traffic networks. On the one hand, Newman fast algorithm is employed to decompose a large-scale traffic network into a set of sub-networks, which greatly narrows the search space and lays the foundation for the following cooperative signal optimization process. On the other hand, a new surrogate-assisted optimizer named RBF-EDA$^2$ is developed by combining a modified RBF model with an advanced variant of EDA called EDA$^2$. Different from the surrogate-assisted optimizer in SACC that needs to simulate a certain percentage of sub-solutions at each generation, RBF-EDA$^2$ conducts signal optimization with EDA$^2$ merely based on the trained RBF model and only the obtained best sub-solution at the end of the optimization will be simulated by the traffic simulator. As a consequence, the simulation requirement of RBF-EDA$^2$ is largely reduced, leading to outstanding signal optimization efficiency. In addition, the employed EDA$^2$ algorithm is capable of capturing the relationships among different signal phases in a sub-network and thus can achieve coordinate signal optimization to some extent.

Benefiting from the above characteristics, SCSO show several advantages over existing traffic signal optimization algorithms. Firstly, the "curse of dimensionality" effect can be alleviated to some extent by decomposing a large-scale traffic network into sub-networks. As a result, SCSO is able to tackle the large-scale traffic signal optimization problem efficiently. Secondly, traffic plans of different sub-networks are optimized separately, which is beneficial to maintain solution diversity and alleviate premature convergence. Thirdly, although the large-scale traffic network has been decomposed, the resulting sub-networks are still closely connected with each other. The traffic flow in a sub-network has widespread impact on its neighborhood sub-networks. SCSO inserts a sub-solution into the current best complete solution and achieves evaluation by simulating the resultant complete solution in the entire traffic network. In this way, the influences of other sub-networks to the current one are naturally embodied in the simulation process, which is helpful to realize information exchange and signal coordination among sub-networks. SCSO updates the current best complete solution step-by-step with the obtained best sub-solution of each sub-network, ultimately providing a high-quality signal plan for the entire traffic network.

## 4. Case study
### 4.1 Experimental setup

A simplified traffic network generated based on real traffic network in an area of Guiyang City in Guizhou Province, China is employed in this study. As shown in Fig. 4, there are totally 43 signalized junctions (J1-J43) in this traffic network, including 26 crossroads and 17 T-junctions. Road 1 to Road 15 (R1-R15) are both the entrances and exits of this network.

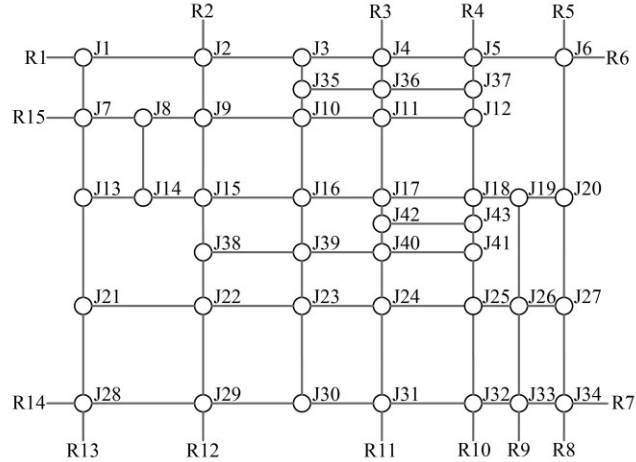

Figure 4. Simplified traffic network in an area of Guiyang City.

All the experiments are implemented in a popular integrated traffic simulation platform [33,50] specially designed for traffic signal optimization. In this simulation platform, MATLAB is employed to execute optimization algorithm and produce candidate solutions (signal plans), and a modern microscopic traffic simulator is used to simulate the traffic dynamics and evaluate the effectiveness of signal plans. Crossroads and T-junctions in this traffic network are designated with different signal phases as explained in Section 3.1. The duration time of each signal phase is set within $[20, 50] \in \mathbf{Z}^+$. Vehicles get into the traffic network from the fifteen entrances and travel in the network randomly according to the predefined tuning rates with a maximum speed of 50 km/h. Table 2 shows the traffic input, in which Case 1 and Case 2 denotes the input traffic volumes in morning peak hour and evening peak hour, respectively. The turning rates of all junctions in the traffic network are presented in Table 3, in which LT, RT and TH denotes the left-turn, right-turn and through ratio, respectively. The simulation time for each candidate signal plan is set to 500 seconds and the average delay time of vehicles at each junction is collected by the traffic simulator as fitness value. To ensure the fairness of comparison, 25 independent runs are conducted for each algorithm and the allowed maximum number of traffic simulations for each run is set to 5000. Although the above experimental environment cannot perfectly simulate real traffic situations, it is still very meaningful to verify the efficiency of SCSO. Similar experimental setup method was also widely used in many existing works [7,33,40].

Table 2. Input traffic volumes (vehicle per hour) in morning peak hour (Case 1) and evening peak hour (Case 2).

|  | R1 | R2 | R3 | R4 | R5 | R6 | R7 | R8 | R9 | R10 | R11 | R12 | R13 | R14 | R15 |
|---|---|---|---|---|---|---|---|---|---|---|---|---|---|---|---|
| Case 1 | 252 | 347 | 497 | 184 | 845 | 180 | 102 | 786 | 159 | 525 | 116 | 253 | 103 | 152 | 926 |
| Case 2 | 327 | 361 | 426 | 135 | 543 | 142 | 39 | 561 | 146 | 617 | 69 | 696 | 250 | 221 | 723 |

Table 3. Turning rates of all junctions in the traffic network.

|  | Southbound | | | Westbound | | | Northbound | | | Eastbound | | |
| --- | --- | --- | --- | --- | --- | --- | --- | --- | --- | --- | --- | --- |
|  | LT | TH | RT | LT | TH | RT | LT | TH | RT | LT | TH | RT |
| J1 | - | - | - | 0.49 | 0.51 | - | 0.20 | - | 0.80 | - | 0.67 | 0.33 |
| J2 | 0.40 | 0.55 | 0.05 | 0.60 | 0.30 | 0.10 | 0.05 | 0.41 | 0.54 | 0.08 | 0.52 | 0.40 |
| J3 | - | - | - | 0.67 | 0.33 | - | 0.30 | - | 0.70 | - | 0.44 | 0.56 |
| J4 | 0.25 | 0.60 | 0.15 | 0.56 | 0.29 | 0.15 | 0.19 | 0.50 | 0.31 | 0.14 | 0.33 | 0.53 |
| J5 | 0.09 | 0.80 | 0.11 | 0.72 | 0.20 | 0.08 | 0.18 | 0.70 | 0.12 | 0.10 | 0.22 | 0.68 |
| J6 | 0.08 | 0.71 | 0.21 | 0.70 | 0.24 | 0.06 | 0.19 | 0.61 | 0.20 | 0.14 | 0.48 | 0.38 |
| J7 | 0.68 | 0.28 | 0.04 | 0.26 | 0.44 | 0.30 | 0.08 | 0.38 | 0.54 | 0.24 | 0.59 | 0.17 |
| J8 | - | - | - | 0.64 | 0.36 | - | 0.22 | - | 0.78 | - | 0.56 | 0.44 |
| J9 | 0.61 | 0.32 | 0.07 | 0.06 | 0.57 | 0.37 | 0.08 | 0.30 | 0.62 | 0.21 | 0.70 | 0.09 |
| J10 | 0.52 | 0.43 | 0.05 | 0.40 | 0.38 | 0.22 | 0.06 | 0.38 | 0.56 | 0.17 | 0.51 | 0.31 |
| J11 | 0.54 | 0.34 | 0.12 | 0.33 | 0.50 | 0.17 | 0.13 | 0.29 | 0.58 | 0.14 | 0.60 | 0.26 |
| J12 | - | 0.93 | 0.07 | - | - | - | 0.05 | 0.95 | - | 0.74 | - | 0.26 |
| J13 | 0.61 | 0.39 | - | 0.58 | 0.42 | - | - | 0.28 | 0.72 | - | - | - |
| J14 | 0.68 | - | 0.32 | - | 0.19 | 0.81 | - | - | - | 0.68 | 0.32 | - |
| J15 | 0.16 | 0.53 | 0.31 | 0.17 | 0.32 | 0.51 | 0.26 | 0.61 | 0.13 | 0.54 | 0.27 | 0.19 |
| J16 | 0.08 | 0.88 | 0.04 | 0.30 | 0.33 | 0.37 | 0.04 | 0.89 | 0.07 | 0.30 | 0.45 | 0.25 |
| J17 | 0.08 | 0.82 | 0.10 | 0.43 | 0.34 | 0.23 | 0.10 | 0.79 | 0.11 | 0.23 | 0.21 | 0.56 |
| J18 | 0.08 | 0.88 | 0.04 | 0.60 | 0.28 | 0.12 | 0.07 | 0.80 | 0.13 | 0.10 | 0.24 | 0.66 |
| J19 | - | - | - | 0.38 | 0.62 | - | 0.37 | - | 0.63 | - | 0.58 | 0.42 |
| J20 | - | 0.88 | 0.12 | - | - | - | 0.21 | 0.79 | - | 0.17 | - | 0.83 |
| J21 | 0.78 | 0.22 | - | 0.41 | - | 0.59 | - | 0.28 | 0.72 | - | - | - |
| J22 | 0.47 | 0.33 | 0.21 | 0.09 | 0.70 | 0.21 | 0.19 | 0.30 | 0.51 | 0.21 | 0.67 | 0.12 |
| J23 | 0.14 | 0.71 | 0.15 | 0.23 | 0.16 | 0.61 | 0.12 | 0.77 | 0.11 | 0.41 | 0.35 | 0.24 |
| J24 | 0.32 | 0.63 | 0.05 | 0.64 | 0.23 | 0.13 | 0.06 | 0.52 | 0.42 | 0.12 | 0.33 | 0.55 |
| J25 | 0.30 | 0.55 | 0.15 | 0.25 | 0.42 | 0.33 | 0.16 | 0.50 | 0.34 | 0.34 | 0.45 | 0.21 |
| J26 | 0.33 | 0.51 | 0.16 | 0.11 | 0.79 | 0.10 | 0.16 | 0.31 | 0.53 | 0.07 | 0.85 | 0.08 |
| J27 | - | 0.72 | 0.28 | - | - | - | 0.25 | 0.75 | - | 0.56 | - | 0.44 |
| J28 | 0.25 | 0.62 | 0.13 | 0.02 | 0.13 | 0.85 | 0.05 | 0.80 | 0.15 | 0.80 | 0.18 | 0.02 |
| J29 | 0.26 | 0.62 | 0.12 | 0.14 | 0.37 | 0.49 | 0.18 | 0.55 | 0.27 | 0.21 | 0.66 | 0.13 |
| J30 | 0.38 | - | 0.62 | - | 0.66 | 0.34 | - | - | - | 0.59 | 0.41 | - |
| J31 | 0.33 | 0.47 | 0.20 | 0.18 | 0.57 | 0.25 | 0.28 | 0.50 | 0.22 | 0.22 | 0.61 | 0.17 |
| J32 | 0.23 | 0.60 | 0.17 | 0.12 | 0.28 | 0.60 | 0.16 | 0.60 | 0.24 | 0.63 | 0.26 | 0.11 |
| J33 | 0.19 | 0.55 | 0.26 | 0.05 | 0.84 | 0.11 | 0.21 | 0.64 | 0.15 | 0.12 | 0.82 | 0.06 |
| J34 | 0.19 | 0.45 | 0.36 | 0.19 | 0.40 | 0.41 | 0.26 | 0.56 | 0.18 | 0.50 | 0.34 | 0.16 |
| J35 | 0.18 | 0.82 | - | 0.73 | - | 0.27 | - | 0.80 | 0.20 | - | - | - |
| J36 | 0.22 | 0.47 | 0.31 | 0.11 | 0.34 | 0.54 | 0.24 | 0.58 | 0.18 | 0.56 | 0.32 | 0.12 |
| J37 | - | 0.94 | 0.06 | - | - | - | 0.10 | 0.90 | - | 0.68 | - | 0.32 |
| J38 | 0.43 | 0.57 | - | 0.37 | - | 0.63 | - | 0.59 | 0.41 | - | - | - |
| J39 | 0.24 | 0.60 | 0.16 | 0.41 | 0.23 | 0.36 | 0.06 | 0.59 | 0.35 | 0.32 | 0.32 | 0.36 |
| J40 | 0.39 | 0.45 | 0.16 | 0.21 | 0.35 | 0.44 | 0.14 | 0.51 | 0.35 | 0.40 | 0.41 | 0.19 |
| J41 | - | 0.93 | 0.07 | - | - | - | 0.09 | 0.91 | - | 0.78 | - | 0.22 |
| J42 | 0.34 | 0.66 | - | 0.38 | - | 0.62 | - | 0.70 | 0.30 | - | - | - |
| J43 | - | 0.49 | 0.51 | - | - | - | 0.36 | 0.64 | - | 0.56 | - | 0.44 |

**4.2 Comparison with existing algorithms**

To verify the efficiency of SCSO, we compare it with six mate-heuristic algorithms for fixed-time signal optimization, including ABC, DHS and Jaya [3], PSO [7], TSO [4] and mEDA$_{ve}$ [33]. These algorithms have all been briefly introduced in Section 2.1. Among them, ABC, DHS, Jaya and PSO have been developed and employed to achieve fixed-time signal optimization in large-scale traffic networks. TSO and mEDA$_{ve}$ were reported to perform well on medium-scale traffic networks, they are also included in our experiment to make the comparison more comprehensive. The number of co-evolution cycles of SCSO is set as $c = 2$, the parameters of the other algorithms are all set to the default values given in their original papers to ensure the fairness of comparison.

Figures. 5(a) and (b) present the evolution histories of SCSO and its six competitors on Case 1 and Case 2, respectively. From Fig. 5, the following comments can be made:

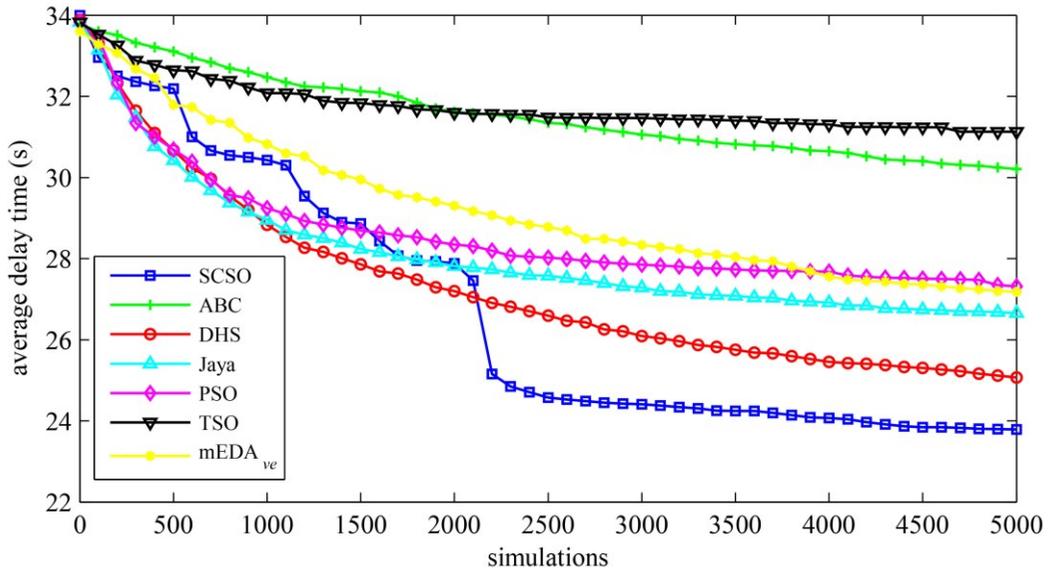

(a)

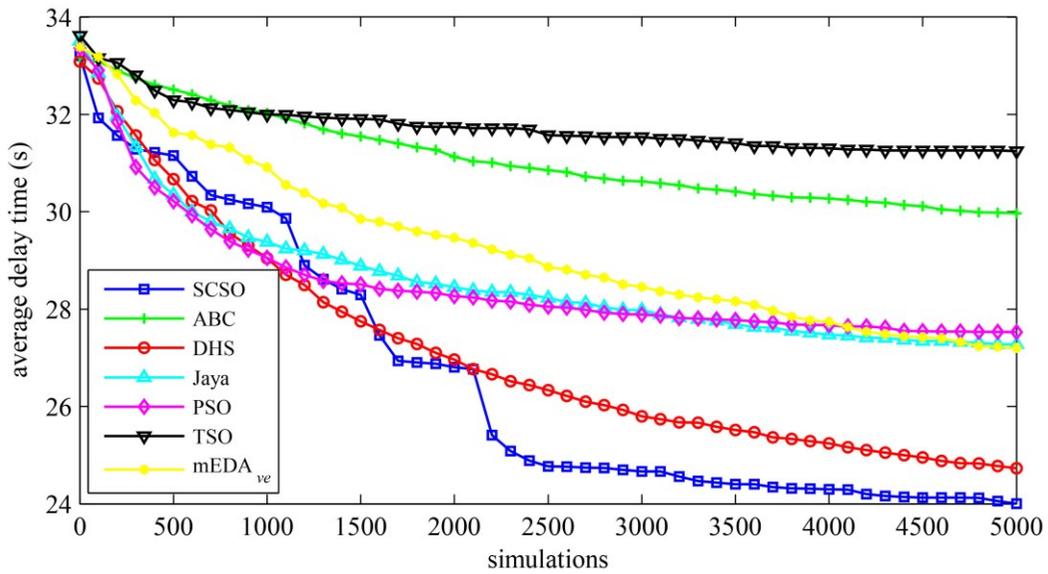

(b)

Figure 5. Evolution history of SCSO and its six competitors on (a) Case 1 and (b) Case 2.

1) In terms of solution quality, SCSO successfully obtains the best final results on both test cases and performs significantly better than all the other algorithms. The signal plan found by SCSO can greatly reduce the average delay time of vehicles in the large-scale network, demonstrating the powerful optimization ability of SCSO. DHS achieves the second best performance and the final results obtained by it are relatively close to that of SCSO. The performances of Jaya, PSO and mEDA$_{ve}$ are similar to each other, ABC and TSO perform poorly on both cases. What is also worth noticing is that mEDA$_{ve}$ and TSO are originally designed for signal optimization in medium-scale traffic networks and the experimental results show that mEDA$_{ve}$ has relatively good scalability. The performance of SCSO and mEDA$_{ve}$ evidences that EDAs might also be a type of effective optimization algorithm for traffic signal optimization.

2) With respect to the convergence performance, DHS, Jaya and PSO exhibit similar and fastest convergence rate at the beginning of the evolution. SCSO shows different convergence rates in different stages because of that it sequentially optimizes the signal plan of different sub-networks. At the middle stage of the evolution, the convergence rate of SCSO is largely increased and exceeds that of other algorithms. On the whole, SCSO could demonstrate the average fastest convergence rate in the first half stage of the evolution. Note that SCSO divides the whole evolution process into two co-evolution cycles, the experimental results indicate that SCSO could quickly locate high-quality solution in the first co-evolution cycle. As for the second half stage of the evolution, the convergence rates of all the seven algorithms are close to each other and are very slow. Reason for the slow convergence rate of SCSO, DHS, Jaya, PSO and mEDA$_{ve}$ mainly lies in that they have found relatively good solutions and are striving to further improve the solution quality. As for TSO and ABC, they always keep a slow convergence rate in the whole optimization process, leading to undesirable performance.

To measure the concrete performance difference between SCSO and the other algorithms, Wilcoxon's rank sum test at a 0.05 significance level is conducted based on the their final optimization results. Table 4 reports the Wilcoxon's rank sum test results (*p*-value) between SCSO and each of its competitors for the two test cases obtained by MATLAB. It is clearly that SCSO performs significantly better than the other six algorithms.

Table 4. Wilcoxon's rank sum test results (*p*-value) between SCSO and each of its competitors for two test cases .

|        | ABC      | DHS      | Jaya     | PSO      | TSO      | mEDAve   |
|--------|----------|----------|----------|----------|----------|----------|
| Case 1 | 1.41E-09 | 8.55E-08 | 1.60E-09 | 5.21E-09 | 1.42E-09 | 1.42E-09 |
| Case 2 | 1.42E-09 | 8.19E-05 | 1.42E-09 | 2.90E-09 | 1.42E-09 | 1.42E-09 |

The appealing performance of SCSO mainly benefits from three aspects. First, SCSO greatly reduces the search space by decomposing a large-scale network into sub-networks. Second, the RBF-EDA$^2$ surrogate-assisted optimizer can efficiently optimize the signal plan of sub-networks using only a few number of simulations. Traffic signals in different sub-networks are separately optimized with different sets of sub-solutions, which contributes to improve the overall solution diversity and avoid premature convergence. Finally, the cooperation among sub-networks in the solution simulation process helps SCSO to find global better signal plans.

### 4.3 Influences of parameters

There are a few parameters in the proposed SCSO, including the number of co-evolution cycles $c$ and the parameters contained in the RBF-EDA$^2$ optimizer. The parameter setting of RBF-EDA$^2$ has been discussed and given in Section 3.2.2.

This section devotes to further analyzing the influence of $c$. After decomposing the original large-scale traffic network into sub-networks, $c$ directly determines number of co-evolution cycles and the number of traffic simulations allocated to each sub-network in each co-evolution cycle as explained in Section 3.2.3. With a small value of $c$, more simulations could be allocated to each sub-network in a co-evolution cycle, which is helpful to achieve fine search of a sub-network. However, fewer co-evolution cycles will also deteriorate the sub-network cooperation ability of SCSO. For instance when $c = 1$, each sub-network will only be optimized once. If the optimizer obtains an undesirable signal plan for a sub-network, there is no second chance to make further improvement. The undesirable signal plan of a sub-network will also have negative influence on the optimization process of other sub-networks, leading to unsatisfying final result. On the contrary, a large value of $c$ can increase the number of co-evolution cycles and the cooperation frequency of sub-networks, but the number of simulations allocated to each sub-network in each co-evolution cycle is reduced.

To investigate the influence of $c$, the performance of SCSO is tested on Case 1 with different values of $c \in \{1, 2, 3, 4\}$. The optimization results obtained by SCSO are shown in Fig. 6. From Fig. 6, two main observations could be made: 1) when $c$ increases from 1 to 4, the final results obtained by SCSO only have slight differences, which indicates that SCSO is robust to the change of $c$. 2) SCSO exhibits the best optimization performance when $c = 2$. More concretely, SCSO tends to obtain better result when $c$ varies from 1 to 2. But it shows certain performance deterioration with the further increase of $c$.

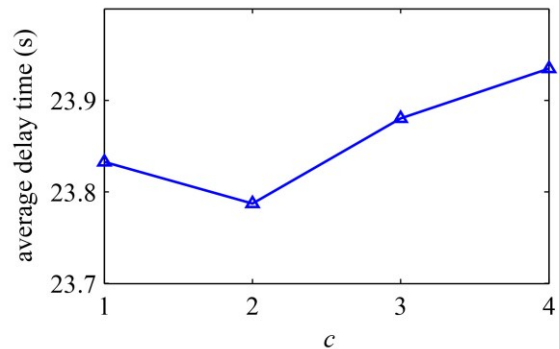

Figure 6. Optimization results obtained by SCSO with different values of $c$ on Case 1.

To delve into the influence of $c$ on the evolution process of SCSO, the evolution histories of SCSO with different $c$ values are maintained and presented in Fig. 7. It can be clearly seen from Fig. 7 that $c$ has great influence on the convergence performance of SCSO. Larger value of $c$ generally leads to faster convergence speed, especially in the earlier stage. This is understandable since a larger value of $c$ means SCSO can quickly accomplish the first co-evolution cycle to locate a relatively good solution. While in the latter stage, the influence of $c$ on the convergence rate of SCSO is almost negligible no matter what the value of $c$ is. The inner reason may lie in that SCSO has converged to high-quality solutions and it is hard to find better ones. Based on the results in Figs. 6 and 7, we suggest setting $c = 2$ to pursue both high-quality solution and fast convergence rate.

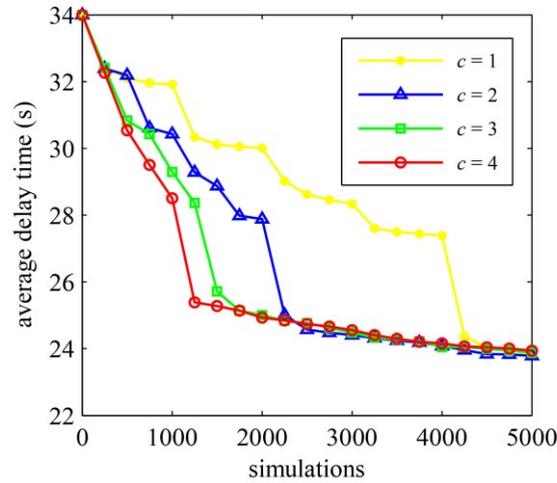

Figure 7. Evolution history of SCSO with different values of $c$ on Case 1.

### 4.4 Robustness to network decomposition result

Traffic network decomposition provides the basis for the cooperative sub-network signal optimization process. To investigate the robustness of SCSO to network decomposition result, the performance of SCSO is tested with two different network decomposition results in this sub-section. In addition to the decomposition result obtained by the Newman fast algorithm, another decomposition result is also generated based on authors' personal knowledge. Figures. 8(a) and (b) present the network decomposition results obtained by the Newman fast algorithm and the personal knowledge-based method, respectively. Junctions in different colors represent different sub-networks.

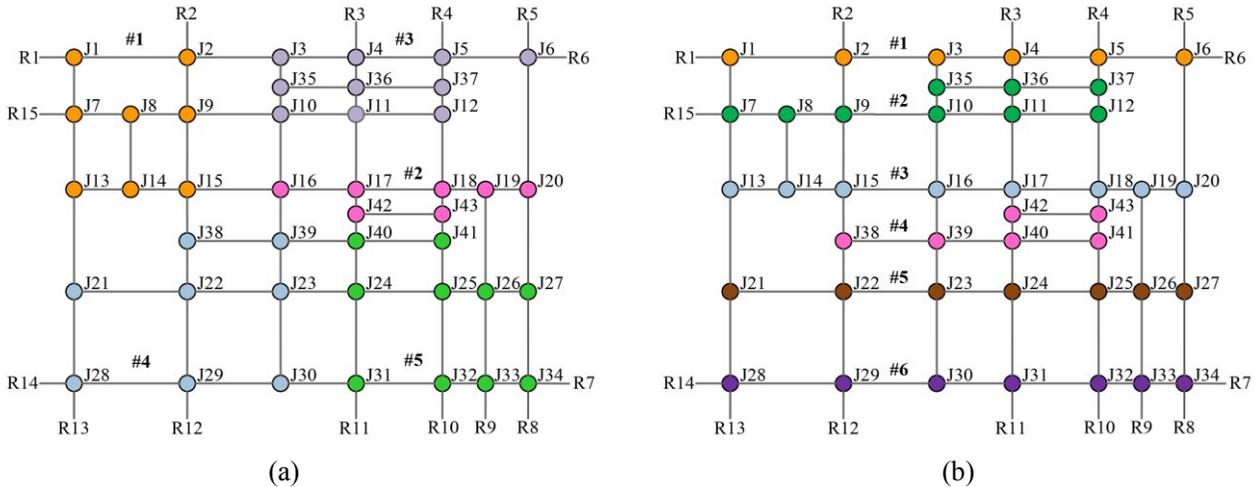

Figure 8. Network decomposition result obtained by (a) Newman fast algorithm and (b) personal knowledge-based method. #1~#5 and #1~#6 denote the sub-network number.

It can be seen from Figs. 8(a) and (b) that the two network decomposition methods divides the original traffic network into five and six sub-networks, respectively. The Newman fast algorithm tends to group close-connected junctions into the same sub-network, while the sub-networks generated based on personal knowledge are generally arterial roads. Combined with the two network decomposition results, two variants of SCSO can be developed. To distinguish them, SCSO with Newman fast algorithm and personal knowledge-based method are denoted as $SCSO_{Newman}$ and $SCSO_{PK}$, respectively. Note that the only difference between the two SCSO variants is the employed network decomposition result, the other algorithm components of them are exactly the same.

Figure 9 shows the evolution histories of $SCSO_{Newman}$ and $SCSO_{PK}$. We can see from Fig. 9 that the two SCSO variants perform similarly in the whole optimization process. $SCSO_{PK}$ converges faster than $SCSO_{Newman}$ in the earlier evolution stage, but is surpassed by $SCSO_{Newman}$ in the latter evolution stage. As a result, $SCSO_{Newman}$ obtains a slightly better final result at the end of the evolution. The observation in Fig. 9 indicates that SCSO has good robustness to different network decomposition results and a proper network decomposition method can improve the performance of SCSO to a certain extent.

To look into the optimization process of each sub-network in each co-evolution cycle, Figs. 10(a) and (b) present the delay reduction (fitness improvement) obtained in each sub-network in the two co-evolution cycles of $SCSO_{Newman}$ and $SCSO_{PK}$ on Case 1, respectively. $SCSO_{Newman}$ and $SCSO_{PK}$ involve different number of sub-networks with different characteristics as presented in Fig. 8. It can be observed from Figs. 10(a) and (b) that the delay reductions obtained in different sub-networks of both algorithms differ greatly, which means that different sub-networks play different roles in affecting the performance of the entire traffic network. Besides, the delay reduction obtained in the first co-evolution cycle is much larger than that obtained in the second co-evolution cycle. That is to say, SCSO can quickly find high-quality solution in the first co-evolution cycle and is also able to gradually improve the solution quality in the second co-evolution cycle. This observation is consistent with the experimental results shown in Fig. 9.

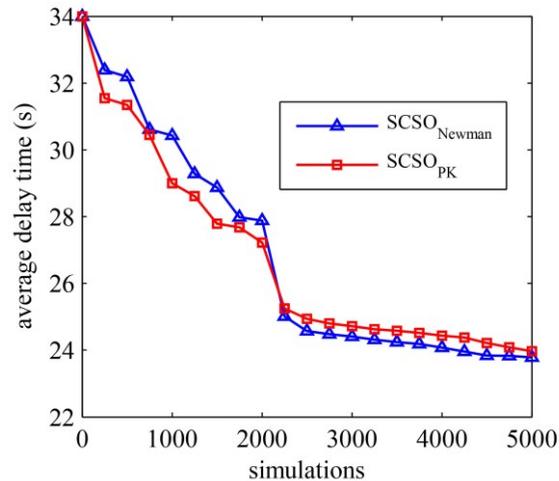

Figure 9. Evolution history of $SCSO_{Newman}$ and $SCSO_{PK}$ on Case 1.

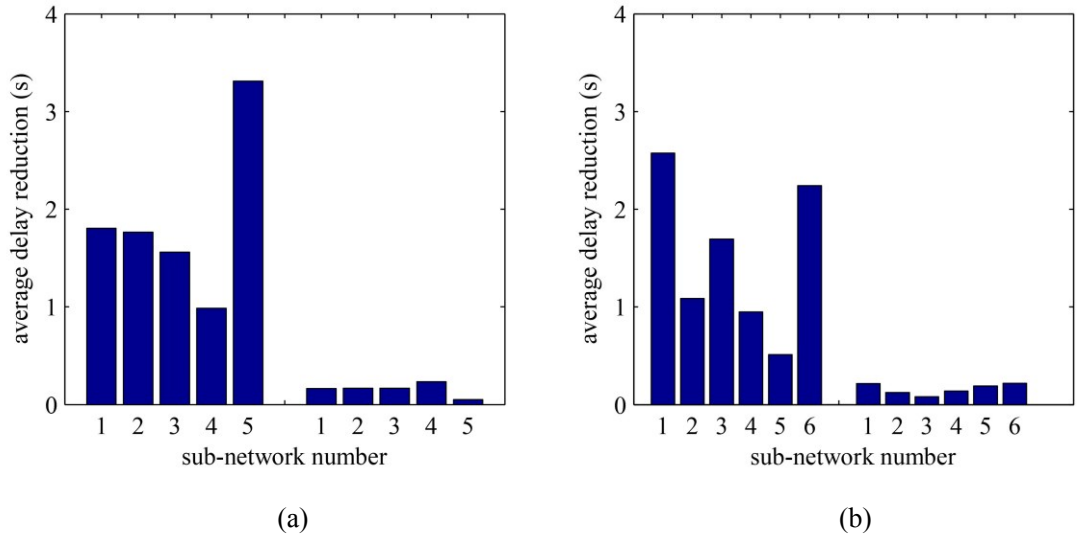

(a)                          (b)

Figure 10. Delay reduction (fitness improvement) obtained in each sub-network in the two co-evolution cycles of (a) SCSO$_{Newman}$ and (b) SCSO$_{PK}$ on Case 1.

### 4.5 Runtime analysis

The RBF-EDA$^2$ optimizer in SCSO can reduce the requirement of expensive traffic simulations, but it needs to train RBF model for sub-networks and conduct optimization with EDA$^2$, bringing extra computational burden. To measure the computational cost of these operations, a runtime analysis is conducted on a Windows 7 computer with CPU @ 3.6GHz.

Figure 11 reports the average computation time for a single run of SCSO, including the total computation time, the computation time spent on traffic simulations and the time spent on all the other algorithmic operations. As we can see from Fig. 11, a single run of SCSO takes over 12 hours and the traffic simulations cost nearly 10 hours, verifying the computationally expensive feature of traffic simulation. By comparison, the computation time spent on all the other operations, including surrogate model training and optimization, becomes less significant. Overall, the RBF-EDA$^2$ optimizer can efficiently improve the signal optimization efficiency of sub-networks without adding too much computational burden.

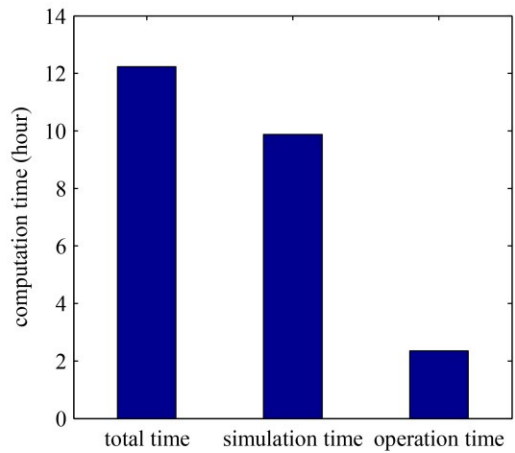

Figure 11. Average computation time for a single run of SCSO.

## 5. Conclusions and future works

In this paper, a surrogate-assisted cooperative signal optimization (SCSO) algorithm is proposed for fixed-time traffic signal optimization in large-scale traffic networks with the goal of reducing the average delay time of vehicles. Different from existing methods that directly tackle the whole traffic network with mate-heuristic optimization algorithms, SCSO first decomposes a large-scale traffic network into a set of smaller sub-networks using Newman fast algorithm, and then employs a surrogate-assisted optimizer named RBF-EDA$^2$ to efficiently optimize the signal plan of sub-networks. The use of RBF model significantly reduces the requirement of expensive traffic simulations and the adopted EDA$^2$ algorithm indicates strong ability in finding high-quality signal plans. Different sub-networks cooperate with each other by sharing their current best sub-solutions for traffic simulation in the solution evaluation process. By virtue of the above characteristics, SCSO demonstrates efficient signal optimization ability in large-scale traffic networks with tight computational budget.

The performance of SCSO is comprehensively studied in a large-scale traffic network containing 43 signalized junctions. Compared with six existing algorithms, SCSO achieves the overall best performance in reducing the average delay time of vehicles, indicating great superiority over its competitors in terms of both solution quality and convergence rate. Further analyses show that SCSO is robust to its parameters and can consistently provide high-quality signal timing plans with different network decomposition results, and the proposed RBF-EDA$^2$ optimizer does not increase the computational burden too much.

In future work, we are interested in testing the performance of SCSO in other large-scale traffic networks with different optimization objectives. It is beneficial to combine SCSO with new network decomposition methods. More efficient surrogate models and optimization algorithms could also be introduced into SCSO to further improve its efficiency.

**Acknowledgement**

This work was supported by the National Natural Science Foundation of China (grant number 61873199), the Natural Science Basic Research Plan in Shaanxi Province of China (grant number 2020JM-059), and the Fundamental Research Funds for the Central Universities [grant numbers xzy022019028].